\documentclass[11pt,a4paper]{article}
\usepackage[hyperref]{emnlp2020}
\usepackage{times}
\usepackage{multirow}
\usepackage{multicol}
\usepackage{latexsym}
\usepackage{booktabs, multirow} % for borders and merged ranges
\usepackage{soul}% for underlines
\usepackage{xcolor,colortbl}
\usepackage{changepage,threeparttable} % for wide tables
\usepackage{tabularx,multirow,booktabs,blindtext}
\usepackage{amsmath,array,graphicx}
\usepackage{float}
\usepackage{verbatim}
\usepackage{booktabs}
\usepackage{multirow}
\usepackage{graphicx}

\usepackage{microtype}

\aclfinalcopy

\newcommand{\Secref}[1]{Section~\ref{#1}}
\newcommand{\secref}[1]{Section~\ref{#1}}

\newcommand{\Figref}[1]{Figure~\ref{#1}}

\newcommand{\Tabref}[1]{Table~\ref{#1}}
\newcommand{\tabref}[1]{Table~\ref{#1}}

\newcommand{\BertEn}{\mbox{BERT-En}}
\newcommand{\BertMultilingual}{\mbox{BERT-Multi}}
\newcommand{\TurkishNLI}{\mbox{NLI-TR}}
\newcommand{\TurkishMultiNLI}{\mbox{MultiNLI-TR}}
\newcommand{\FBTR}{\mbox{MultiNLI-TR$^{\textrm{XNLI}}$}}

\newlength{\Oldarrayrulewidth}
\newcommand{\Cline}[2]{%
  \noalign{\global\setlength{\Oldarrayrulewidth}{\arrayrulewidth}}%
  \noalign{\global\setlength{\arrayrulewidth}{#1}}\cline{#2}%
  \noalign{\global\setlength{\arrayrulewidth}{\Oldarrayrulewidth}}}
  
\title{Data and Representation for Turkish Natural Language Inference}

\author{%
Emrah Budur,$^{1,2}$
R{\i}za \"{O}z\c{c}elik,$^{2}$ 
Tunga G\"{u}ng\"{o}r,$^{2}$
and 
Christopher Potts$^{3}$ 
\\[1ex]
${}^{1}$Garanti BBVA Technology
\quad
${}^{2}$Bo\u{g}azi\c{c}i University
\quad
${}^{3}$Stanford University
\\[1ex]
{\tt \{emrah.budur, riza.ozcelik, gungort\}@boun.edu.tr},\\ 
{\tt cgpotts@stanford.edu}
}

\date{}

\begin{document}
\maketitle

\begin{abstract}
Large annotated datasets in NLP are overwhelmingly in English. This is an obstacle to progress in other languages. Unfortunately, obtaining new annotated resources for each task in each language would be prohibitively expensive. At the same time, commercial machine translation systems are now robust. Can we leverage these systems to translate English-language datasets automatically? In this paper, we offer a positive response for natural language inference (NLI) in Turkish. We translated two large English NLI datasets into Turkish and had a team of experts validate their translation quality and fidelity to the original labels. Using these datasets, we address core issues of representation for Turkish NLI. We find that in-language embeddings are essential and that morphological parsing can be avoided where the training set is large. Finally, we show that models trained on our machine-translated datasets are successful on human-translated evaluation sets. We share all code, models, and data publicly.
\end{abstract}
\section{Introduction}\label{sec:intro}

Many tasks in natural language processing have been transformed by the introduction of very large annotated datasets. Prominent examples include paraphrase \citep{ganitkevitch-etal-2013-ppdb}, parsing \citep{nivre-etal-2016-universal}, question answering (\citealp{rajpurkar-etal-2016-squad}), machine translation (MT; \citealp{bojar-etal-2014-findings}), and natural language inference (NLI; \citealp{bowman-etal-2015-large, williams-etal-2018-broad}).

Unfortunately, outside of parsing and MT, these datasets tend to be in English. This is not only an obstacle to progress on other languages, but it also limits the field of NLP itself: English is generally not a representative example of the world's languages when it comes to morphology, syntax, or spelling conventions and other kinds of standardization \citep{Munro:2012}, so it's risky to assume that models and results for English will generalize to other languages.

A natural response to these gaps in our dataset coverage might be to launch new annotation efforts for multiple languages. However, this would likely be prohibitively expensive. For example, based on the costs of SNLI \citep{bowman-etal-2015-large} and MultiNLI \citep{williams-etal-2018-broad}, we estimate that each large dataset for NLI would cost upwards of US\,\$50,000 if created completely from scratch.

At the same time, commercial MT systems have improved dramatically in recent years \cite{wu2016-google-translate, johnson2017google, hieber-2017-sockeye2-amazon-translate, hieber-etal-2018-sockeye, tomasello2019neural, felix2020sockeye2}. They now offer high-quality translations between hundreds of language pairs. This raises the question: can we use these MT systems to translate English-language datasets and use the translated versions to drive more genuinely multilingual development in NLP? In this paper, we offer evidence that the answer is ``yes''. 

Using Amazon Translate, we translated SNLI and MultiNLI from English into Turkish to create the first large Turkish NLI data sets, \TurkishNLI , at a tiny fraction of the cost of creating them from scratch. Turkish is an interesting challenge in this context since it is very different from English, most notably in its very free word order and complex morphology. A word in Turkish bears morpho-syntactic properties in the sense that phrases formed of several words in languages like English can be expressed with a single word form. 

In our validation phase (\secref{sec:datasets}), a team of Turkish--English bilingual speakers assessed the quality of a large sample of the translations in \TurkishNLI. They found the quality to be very high, which suggests that translated datasets can provide a foundation for NLI research on a resource-constrained language, even if it has significantly different characteristics from English. 

We then use these datasets to study the roles of pre-trained language models and morphological parsing in successful NLI systems for Turkish (\secref{sec:experiments}). For these experiments, we fit classifiers on top of pre-trained BERT parameters \citep{devlin-etal-2019-bert} and compare the original BERT-base release, the multilingual BERT embeddings released by the BERT team, and the Turkish BERT (BERTurk) embeddings of \citet{schweter2020berturk}. We find BERTurk to be superior to the others for \TurkishNLI.

Morphological parsing is a natural preprocessing step for Turkish due to its complex morphology. Thus, we assess the use of three morphological parsers as the second case study: Zemberek \citep{Zemberek}, BOUN parser \citep{sak2011resources}, and Turkish Morphology \citep{ozturel2019syntactically}. We find that the parsers help where training data is sparse, but the need for a parser disappears as the training data increases. This is a striking finding: one might expect that Turkish would require morphological parsing given its complex word-formation processes. It might be regarded as welcome news, though, since the parsers are expensive to run. In \secref{sec:parsers}, we report on some new optimizations of existing tools to make the relevant parsing jobs feasible, but we would still like to avoid these steps if possible, and it seems that we can for NLI.

Finally, we investigate how models trained on the machine translated datasets perform on the human translations from
XNLI \citep{conneau-etal-2018-xnli}. We find that machine translated and human translated sentences yield similar results, suggesting that it is safe to apply models trained on machine-translated datasets to human-written sentences.
\section{Related Work} 

Early in the development of textual entailment tasks, \citet{mehdad-etal-2010-towards} argued for multilingual versions of them. This led to subsequent explorations of a variety of techniques, including crowdsourcing translations \citep{negri-mehdad-2010-creating,negri-etal-2011-divide}, relying on parallel corpora to support reasoning across languages \citep{mehdad-etal-2011-using}, and automatically translating datasets using MT systems \citep{mehdad-etal-2010-towards, real2018sick, rodrigues_assin2_2020}. This research informed SemEval tasks in 2012 \citep{negri-etal-2012-semeval} and 2013 \citep{negri-etal-2013-semeval} followed by ASSIN 1 \citep{fonseca2016visao} and 2 \citep{real2020assin} shared tasks exploring the viability of multilingual NLI. From the perspective of present-day NLI models, these datasets are very small, but they could be used productively as challenge problems.

More recently, \citet{conneau-etal-2018-xnli} reinvigorated work on multilingual NLI with their XNLI dataset. XNLI provides expert-translated evaluation sets from English into 14 other languages, including Turkish. Though they are valuable resources to push NLI research beyond English, test sets alone are insufficient for in-language training on target languages, which is likely to lower the performance of the resulting systems.

 Although it was not the main focus of the XNLI effort,
\citet{conneau-etal-2018-xnli} distributed machine translations of MultiNLI into other languages, including Turkish, which we call \FBTR\ in this paper. The translations helped them form a strong baseline for their cross-lingual models, which proved superior in their assessments. However, the quality of the translations is crucial, as the authors note. Our hope for \TurkishNLI\ is that it supports effective in-language training.

XNLI's primary focus on test sets rather than training is justified by a wide body of recent results on cross-lingual transfer learning. Multilingual embeddings (embeddings trained on multilingual corpora) have played an important role in these developments. The BERT team \citep{devlin-etal-2019-bert} released multilingual embeddings and demonstrated their value using XNLI. At the same time, BERT models have been released for a variety of individual languages (see \citealp{Wolf2019HuggingFacesTS}) and specialized domains \citep{alsentzer-etal-2019-publicly,lee2020biobert}. While we might expect the language- and domain-specific embeddings to be superior for the kind of data they were trained on, the multilingual versions might be more efficient in large-scale deployments in diverse environments. Balancing these trade-offs is challenging. Here, we offer some insight into these trade-offs for Turkish.

Turkish is a morphologically-rich language in which new word forms are freely created using suffixation. Several morphological parsers \citep{Zemberek, ozturel2019syntactically, sak-etal-2009-stochastic}
and morphological disambiguation systems \citep{Zemberek,sak2011resources}
have been developed for Turkish. The state-of-the-art morphological analyzers can parse with success rates around 95\%. We use three of these parsers in this work to evaluate the role of morphology in NLI systems (\secref{sec:parsers}).
\section{Creating and Validating \TurkishNLI}\label{sec:datasets}
\begin{table*}[!ht]
\centering
\begin{tabular}{llcc}
& & \textbf{English} & \textbf{Turkish} \\ \hline

\multicolumn{1}{|l|}{}  & \multicolumn{1}{p{2.5cm}|}{\textbf{Premise}} & \multicolumn{1}{p{5cm}|}{Three men are sitting near an orange building with blue trim.} & \multicolumn{1}{p{5cm}|}{\"{U}\c{c} adam mavi s\"{u}slemeli turuncu bir binan{\i}n yan{\i}nda oturuyor.} \\ \Cline{0.8pt}{2-4} 

\multicolumn{1}{|l|}{} & \multicolumn{1}{p{2.5cm}|}{ \textbf{Entailment}} & \multicolumn{1}{p{5cm}|}{ Three males are seated near an orange building with blue trim.} & \multicolumn{1}{p{5cm}|}{ \"{U}\c{c} erkek mavi s\"{u}sl\"{u} turuncu bir binan{\i}n yak{\i}n{\i}nda oturuyor.} \\ \cline{2-4}

\multicolumn{1}{|l|}{} & \multicolumn{1}{p{2.5cm}|}{ \textbf{Contradiction}} & \multicolumn{1}{p{5cm}|}{ Three women are standing near a yellow building with red trim.} & \multicolumn{1}{p{5cm}|}{\"{U}\c{c} kad{\i}n k{\i}rm{\i}z{\i} s\"{u}slemeli sar{\i} bir binan{\i}n yan{\i}nda duruyor.}                                                                              \\ \cline{2-4} 
\multicolumn{1}{|l|}{\multirow{-4}{*}{\textbf{SNLI}}}  & \multicolumn{1}{p{2.5cm}|}{ \textbf{Neutral}} & \multicolumn{1}{p{5cm}|}{Three males are seated near an orange house with blue trim and a blue roof.} & \multicolumn{1}{p{5cm}|}{\"{U}\c{c} erkek mavi s\"{u}sl\"{u} ve mavi \c{c}at{\i}l{\i} turuncu bir evin yak{\i}n{\i}nda oturuyor.} \\ \hline
\end{tabular}
\caption{Sample translations from SNLI into \TurkishNLI. Each premise is associated with a hypothesis from each of the three NLI categories. \Tabref{tab:sample_translations_table_all} in our supplementary materials provides MultiNLI examples.}
\label{tab:sample_translations_table_snli}
\end{table*}

\subsection{English NLI Datasets}\label{sec:en-datasets}

We translated the Stanford Natural Language Inference Corpus (SNLI;  \citealp{bowman-etal-2015-large}) and the Multi-Genre Natural Language Inference Corpus (MultiNLI; \citealp{MultiNLI}) to create labeled NLI datasets for Turkish, \TurkishNLI. 

SNLI contains $\approx$570K semantically related English sentence pairs. The semantic relations are entailment, contradiction, and neutral. The premise sentences for SNLI are image captions from the Flickr30K corpus \citep{young-etal-2014-image}, and the hypothesis sentences were written by crowdworkers. SNLI texts are mostly short and structurally simple. We translated SNLI while respecting the train, development (dev), and test splits.

MultiNLI  comprises $\approx$433K sentence pairs in English, and the pairs have the same semantic relations as SNLI. However, MultiNLI spans a broader range of genres, including travel guides, fiction, dialogue, and journalism. As a result, the texts are generally more complex than SNLI. In addition, MultiNLI contains \emph{matched} and \emph{mismatched} dev and test sets, where the sentences in the former set are from the same sources as the training set, whereas the latter consists of texts from different genres than those found in the training set. We translated the training set and both dev sets for \TurkishNLI.

\subsection{Automatic Translation Effort}
\label{sec:translation}
\begin{table*}[!ht]
\centering
\footnotesize
\setlength{\tabcolsep}{4pt}
\begin{tabular}{ l l *{6}{r} }\toprule
& & \multicolumn{3}{c}{\textbf{English}} & \multicolumn{3}{c}{\textbf{Turkish}} \\ \cmidrule{3-8} 
\textbf{Dataset} & \textbf{Fold} & \textbf{Token Count} & \textbf{\begin{tabular}[r]{@{}r@{}}Vocab Size\\  (Cased)\end{tabular}} & \textbf{\begin{tabular}[r]{@{}r@{}}Vocab Size\\  (Uncased)\end{tabular}} & \textbf{Token Count} & \textbf{\begin{tabular}[r]{@{}r@{}}Vocab Size\\  (Cased)\end{tabular}} & \textbf{\begin{tabular}[r]{@{}r@{}}Vocab Size\\  (Uncased)\end{tabular}} \\ \midrule
\multirow{3}{*}{\textbf{SNLI}}&\textbf{Train} & 5900366 & 38565 & 32696 & 4298183 & 78786 & 66599 \\
&\textbf{Dev}  & 120900 & 6664 & 6224 & 88668 & 11455 & 10176                         \\
&\textbf{Test} & 120776 & 6811 & 6340 & 88533 & 11547 & 10259                         \\ \midrule
\multirow{3}{*}{\textbf{MultiNLI}}&\textbf{Train}  & 6356136 & 81937 & 66082 & 4397213 & 216590 & 187053 \\
&\textbf{Matched Dev}    & 161152 & 14493 & 12659 & 112192 & 27554 & 24872                         \\
&\textbf{Mismatched Dev} & 170692 & 12847 & 11264 & 119691 & 26326  & 23941  \\
\bottomrule
\end{tabular}
\caption{Comparative statistics for the English and Turkish NLI datasets. The Turkish translations have larger vocabularies and lower token counts due to the highly agglutinating morphology of Turkish as compared to English.}
\label{tab:dataset_stats}
\end{table*}
As we noted in \secref{sec:intro}, Turkish is a resource-constrained language with few labeled data sets compared to English. Furthermore, Turkish has a fundamentally different grammar from English that could hinder transfer-learning approaches. These facts motivate our effort to translate SNLI and MultiNLI from English to Turkish. We employ an automatic MT system and hope that it will deliver high-quality translations that we can use for NLI research and system development in Turkish.

We used Amazon Translate, a commercial neural machine translation service. Translation of all folds of SNLI and MultiNLI cost just US\,\$2K (vs.~the $\approx$US\,\$100K we would expect for replicating these two datasets from scratch) and five days with no parallelization. We refer to the translated datasets as SNLI-TR and MultiNLI-TR, and collectively as \TurkishNLI. Translation examples are provided in \Tabref{tab:sample_translations_table_snli}. We publicly share \TurkishNLI.\footnote{\url{https://github.com/boun-tabi/NLI-TR}}

SNLI-TR and MultiNLI-TR are different from SNLI and MultiNLI in terms of token counts and vocabulary sizes. \Tabref{tab:dataset_stats} illustrates these features before and after translation. For each fold in each dataset, translation decreased the number of tokens in the corpus, but it increased the vocabulary sizes drastically, in both the cased and uncased versions. Both of these differences are expected: many multiword expressions in English are translated into individual words due to the agglutinating nature of Turkish. For instance, the four-word English expression ``when in your home'' can be translated to the single word ``evinizdeyken''.

\Tabref{tab:dataset_stats} also reflects the complexity difference between SNLI and MultiNLI that we noted in \secref{sec:en-datasets}. Though SNLI contains more sentence pairs than MultiNLI, it has fewer tokens and a smaller vocabulary.

\subsection{Translation Quality Assurance}
\label{sec:validation}

Two major risks arise when using MT systems to translate NLI datasets. First, the translation quality might be low. Second, even if the individual sentences are translated correctly, the nature of the mapping from the source to the target language might affect the semantic relations between sentences. For example, English has the words ``boy'' and ``girl'' to refer to male and female children, and both those words can be translated to a gender-neutral Turkish word ``\c{c}ocuk''. Now, consider a premise sentence ``A boy is running'' and its contradiction pair ``A girl is running''. Both sentences can be translated fluently into the same Turkish sentence, ``\c{C}ocuk ko\c{s}uyor'', which changes the semantic relation from contradiction to entailment. 

Thus, to determine the viability of \TurkishNLI\ as a tool for NLI research, we must assess both translation quality and the consistency of the NLI labels. To do this, we assembled a team of ten Turkish--English bilingual speakers who were familiar with the NLI task and were either MSc.\ candidates or graduates in a relevant field.

For expert evaluation, we grouped the translations into example sets of four sentences as in \tabref{tab:sample_translations_table_snli}, where the first sentence (premise) is semantically related to the rest (hypotheses). We distributed the sets to the experts so that each set (and sentence) was examined by five randomly chosen experts and each expert co-examined approximately the same number of sets with each other expert. Each expert evaluated the translation by (i) grading the translation quality between 1 and 5 (inclusive; 5 the best) and (ii) checking if the translation altered the semantic relation. We distributed an annotation guide\footnote{\url{https://github.com/boun-tabi/NLI-TR}} to the team to standardize the criteria. In total, 500 example sets (2,000 translated sentences) were examined by five experts, yielding 10,000 annotations.

We use the average translation score of the annotations to
estimate translation quality. For label consistency, there are two comparisons we can make, since we have five new annotations per example. The \emph{annotation-level} analysis compares each new annotation with the gold label on the original English example. The \emph{majority-level} analysis compares only the majority label (if any) of the five new annotations with the English gold label. The annotation-level analysis is more stringent, whereas the majority-level analysis directly connects with how we expect \TurkishNLI\ to be most commonly used. \Tabref{tab:dataset_quality_stats_v2} reports these analyses for SNLI and MultiNLI. The results are extremely reassuring. First, average translation quality is near 5 (ceiling) for all the splits. Second, annotation-level label consistency is over 90\% and majority-level label consistency is over 95\%, indicating that the linguistic differences between English and Turkish are not a major issue for preserving NLI labels.

\begin{table*}[tp]
\centering
\begin{tabular}{llccc}\toprule
\textbf{Dataset} & \textbf{Fold} & 
\textbf{\begin{tabular}[c]{@{}c@{}}Translation Quality\end{tabular}}
 & 
\textbf{\begin{tabular}[c]{@{}c@{}}Annotation-level \\ Label  Consistency \end{tabular}} & \textbf{\begin{tabular}[c]{@{}c@{}}Majority-level \\ Label  Consistency \end{tabular}}\\ \midrule
\multirow{3}{*}{\textbf{SNLI-TR}}&\textbf{Train} & 4.55 (0.78) & 92.62\%  & 98.67\% \\
&\textbf{Dev}  & 4.46 (0.90) & 90.53\% & 95.33\% \\
&\textbf{Test} & 4.45 (0.86) & 87.87\%  & 94.00\% \\ \midrule
\multirow{3}{*}{\textbf{MultiNLI-TR}}
&\textbf{Train}  & 4.56 (0.80) & 89.96\% & 96.22\% \\
&\textbf{Matched Dev}    & 4.42 (0.86) & 88.53\%  & 95.33\% \\
&\textbf{Mismatched Dev} & 4.49 (0.82) & 92.53\% & 98.00\% \\ \midrule
\multirow{1}{*}{\textbf{ }}&\textbf{All} & 4.51 (0.82) & 90.72\%  & 96.73\%\\
\bottomrule
\end{tabular}
\caption{Translation quality and label consistency of the translations in SNLI-TR and MultiNLI-TR  based on expert judgements. For the quality ratings (1--5), we report mean and standard deviation (in parentheses). For label consistency, we report the percentage of labels in SNLI-TR and MultiNLI-TR judged consistent with the original label, both in annotation- and sentence-level.}
\label{tab:dataset_quality_stats_v2}
\end{table*}

To assess the reliability of the translation quality scores, we calculated the Intra-Class Correlation (ICC; \citealt{mcgraw1996forming}). ICC is frequently adopted in medical studies to assess ordinal annotations provided by experts randomly drawn from a team. Its assumptions align well with our evaluation scheme. We obtained an ICC of 0.8426, which suggests excellent agreement \citep{cicchetti1994guidelines,hallgren2012computing}.

We also computed Krippendorff's alpha \citep{krippendorff1970estimating}, which is an inter-annotator agreement metric used more commonly in NLP. This metric is suitable for both nominal and ordinal annotations involving multiple annotators. We calculated intercoder reliability of the ordinally-scaled translation quality score as 0.47. Our annotation-level label consistency yielded a score of 0.78 whereas our majority-level label consistency resulted in a score of 0.99. In contrary to the perfect agreement in the majority-level label consistency, the Krippendorff's alpha values of annotation-level labels and translation quality scores suggest less overall agreement than our ICC values do, but they are still acceptable, and ICC is arguably the more appropriate metric for our study. Krippendorff's alpha is generally used for large, diverse annotation teams, and its penalties for disagreements are known to be harsh. 

Overall, it seems that the very high estimates of translation quality and label consistency of \TurkishNLI\ are trustworthy, and only a small percentage of premise--hypothesis have inconsistent semantic labels between their original and translated forms. Still, we would like to better understand why inconsistencies do arise. To this end, we inspected all 49 label-inconsistent pairs in our annotations. We find that low translation quality is the leading source of such errors, which further emphasizes how essential it is to work with high-quality translations. 

Of the label-inconsistent pairs with good translations, we find that about 20 probably trace to differing perspectives on how to apply the NLI annotation guidelines. Relatedly, \citet{conneau-etal-2018-xnli} find that NLI labels often cannot be completely recovered by different annotators even with no sentence modifications.

Finally, we did find one example of label inconsistency that traces to a subtle difference between the English and Turkish lexicons. In this example,  the premise ``Your speeches are inflammatory'' was translated to Turkish as ``Konu\c{s}malar{\i}n{\i}z \c{c}ok k{\i}\c{s}k{\i}rt{\i}c{\i}'', which can be back-translated as ``Your speeches are provocative'', while its entailment hypothesis ``Your speeches upset people" was translated as ``Konu\c{s}malar{\i}n insanlar{\i} \"{u}z\"{u}yor", equivalent to ``Your speeches make people sad''. Annotators agreed that both of these translations are of maximum quality, but also stated that the Turkish pair should be labeled neutral. As bilingual speakers, we feel that this is essentially correct; the relevant English and Turkish adjectives are subtly different in ways that affect the NLI label. However, such examples seem to be rare and so pose minimal risk for conducting research using \TurkishNLI. 

\section{Experiments}\label{sec:experiments}

\subsection{Case Study I: Comparing BERT models on Turkish NLI Datasets}\label{sec:case-study-i}

The arrival of pre-trained model-sharing hubs (e.g., Tensorflow Hub,\footnote{\url{https://github.com/tensorflow/hub}} PyTorch Hub,\footnote{\url{https://pytorch.org/hub}} and Hugging Face Hub\footnote{\url{https://huggingface.co/models}}) has democratized access to Transformer-based models \citep{Vaswani-et-al-2017}, which are mostly in English. Combined with the abundance of labeled English datasets for fine-tuning, this has increased the performance gap between English and resource-constrained languages.

Here, we use \TurkishNLI\ to analyze the effects of pretraining Transformer-based models. We compare three BERT models trained on different corpora by fine-tuning them on \TurkishNLI. The results quantify the importance of having high-quality, language-specific resources.

\subsubsection{Experimental Settings}

We compared cased BERT-English (\BertEn), \BertMultilingual, and BERTurk \citep{schweter2020berturk}. \BertEn\ is the original BERT-base model released by \citet{devlin-etal-2019-bert}, which used an English-only corpus for training. \BertMultilingual\ was released by the BERT team as well, and was trained on a corpus containing texts from 104 languages, including Turkish. \citeauthor{schweter2020berturk}'s BERTurk also uses the same model architecture and is trained on a Turkish corpus ($\approx$30GB).

We fine-tuned each model on train folds of \TurkishNLI\ separately and fixed the maximum sequence length to 128 for all experiments. Similarly, we used a common learning rate of $2\times 10^{-5}$ and batch size of 8 with no gradient accumulation. We fine-tuned each model for 3 epochs using HuggingFace's Transformers Library \cite{Wolf2019HuggingFacesTS}. We evaluated the models on the test set of SNLI-TR and the \emph{matched} and \emph{mismatched} dev splits of MultiNLI-TR. \Tabref{tab:case_study_i_results_table} reports the accuracy of each model on the evaluation sets.

\begin{table*}[!ht]\centering
\setlength{\tabcolsep}{12pt}
\begin{tabular}{l c c c c}
\toprule
\textbf{}  &\multicolumn{2}{c}{\textbf{SNLI-TR}} &\multicolumn{2}{c}{\textbf{MultiNLI-TR}} \\\cmidrule{2-5}
\textbf{Model Name} & \textbf{Dev} & \textbf{Test} & \textbf{Matched Dev} &\textbf{Mismatched Dev} \\\midrule
\textbf{\BertEn} & 81.83\% &82.09\% &69.98\% &70.56\% \\
\textbf{\BertMultilingual} & 85.37\%&85.12\% &75.97\% &76.34\% \\
\textbf{BERTurk}& \textbf{87.28\%} &\textbf{87.04\%} &\textbf{79.58\%} &\textbf{80.87\%} \\
\bottomrule
\end{tabular}
\caption{Accuracy results for the publicly available cased BERT models on \TurkishNLI. BERTurk performed the best in all three evaluations, highlighting the value of language-specific resources for NLI.}
\label{tab:case_study_i_results_table}
\end{table*}

\subsubsection{Results}

\Tabref{tab:case_study_i_results_table} demonstrates that \TurkishNLI\ can be used to train high quality Turkish NLI models. We observe that every model performed better on the dev and test folds of SNLI-TR than the dev folds of MultiNLI-TR, which is an expected outcome given the greater complexity of MultiNLI compared to SNLI. The translation effort seems to have preserved this fundamental difference between the two datasets. 

In addition, BERTurk, which was trained on a Turkish corpus, achieved the highest accuracy, and \BertMultilingual, which used a smaller Turkish corpus, was ranked the second, consistently on every evaluation fold. The ranking emphasizes the importance of having a Turkish corpus for pre-training.

\subsection{Case Study II: Comparing Morphological Parsers on Turkish NLI Datasets}\label{sec:parsers}

%Turkish is an agglutinative language in which suffixes are commonly cascaded to create new words that attributes importance to morphological parsing for many applications. 
In this case study, we use \TurkishNLI\ to compare three morphological parsers with regular tokenization. We train a BERT model from scratch utilizing each approach for pretraining
% from \secref{sec:case-study-i}
and use \TurkishNLI\ for fine-tuning. This leads to the striking result that morphology adds additional information where training data is sparse, but its importance shrinks as the dataset grows larger.

\subsubsection{Experimental Settings}

\paragraph{Morphological Parsers}
We use Zemberek \citep{Zemberek}, BOUN \citep{sak2011resources}, and Turkish Morphology \citep{ozturel2019syntactically} as parsers and compare them with an approach that does not do morphological parsing. 

Zemberek is a mainstream Turkish NLP library used in research \citep{Buyuk2020context,Kuyumcu2019,Ozer2018diacritic,can2017unsupervised, dehkharghani2016sentiturknet,Gulcehre2015using} and applications such as iOS 12.2 and Open Office. It has 67,755 entries in its lexicon and uses a rule-based parser. BOUN implements the Turkish morphology rules described by \citet{oflazer1994two} with a Finite State Transducer, and its lexicon has 55,278 entries. Finally, Turkish Morphology is an OpenFST-based \citep{OpenFst} morphological parser that was recently released by Google Research and uses a lexicon with 47,202 entries.

Out of the box, Zemberek and BOUN can parse 398K and 51K tokens per minute respectively, whereas Turkish Morphology can process only 1K tokens. We sped up Turkish Morphology to parse 11 times more tokens per minute by implementing a dynamic programming wrapper \citep{bellman1952DynamicProgramming} that increased the cache hit ratio to 89.9\%. This technique is already used by Zemberek.

\paragraph{Pretraining}

\begin{table*}[tp]
\centering
\setlength{\tabcolsep}{12pt}
\begin{tabular}{l c c c c}\toprule
\textbf{}  &\multicolumn{2}{c}{\textbf{SNLI-TR}} &\multicolumn{2}{c}{\textbf{MultiNLI-TR}} \\\cmidrule{2-5}
&\textbf{Dev} & \textbf{Test} & \textbf{\begin{tabular}[l]{@{}l@{}}Matched   Dev\end{tabular}} &\textbf{\begin{tabular}[l]{@{}l@{}}Mismatched   Dev\end{tabular}} \\
\midrule
\textbf{No Parser} & 76.50\% & 76.59\% &58.24\% &60.01\% \\
\textbf{Zemberek} & 76.47\% & 76.71\% &59.01\% &60.44\% \\
\textbf{BOUN Parser} &\textbf{76.64}\% &\textbf{76.89}\% &59.99\% &61.29\% \\
\textbf{Turkish Morphology} & 76.00\%  & 76.36\% &\textbf{60.13}\% &\textbf{62.00}\% \\ 
\bottomrule
\end{tabular}
\centering
\caption{Accuracy results for different morphology approaches on \TurkishNLI. To facilitate running many experiments, these results are for pretraining on just 
one-tenth of the Turkish corpus used by BERTurk and fine-tuning on \TurkishNLI\ for just one epoch.}
\label{tab:case_study_ii_results}
\end{table*}

To conduct a wide range of experiments on a limited budget, we opted to use one-tenth ($\approx$4GB, 500M tokens) of the Turkish corpus used by BERTurk \citep{schweter2020berturk} to pretrain BERT models. We analyzed each token morphologically using Zemberek, BOUN, and Turkish Morphology and trained a BERT model using the stems of the tokens only. For the model that does not utilize morphological information, we used tokens as they are. We used the \texttt{BertWordPieceTokenizer} class of HuggingFace Tokenizers\footnote{\url{https://github.com/huggingface/tokenizers}} with the same set of parameters for each model.

We trained each model on a single Tesla V100 GPU of an NVIDIA DGX-1 system, allocating 128GB memory for 1 day. We split the dataset into 30 equal shards for parallel processing, where each shard comprises 1M sentences, and  shuffled the shards prior to training to reduce the adverse effects of variance across the sentence styles in the different shards \cite{Goodfellow-et-al-2016}. We used an effective batch size of 128 with gradient accumulation to address memory limitations. 

\begin{figure*}[!ht]
    \centering
    \includegraphics[width=\textwidth]{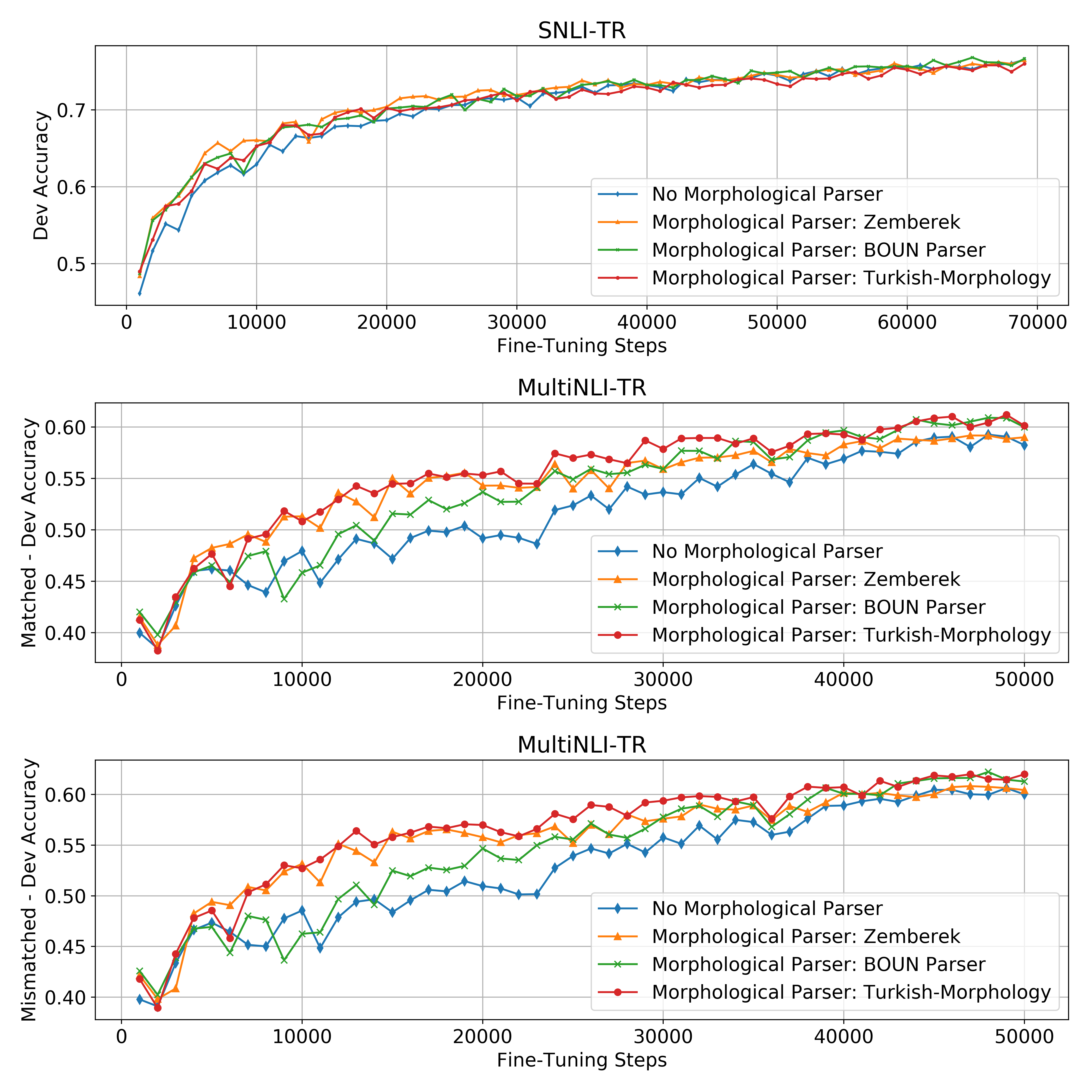}
    
    \vspace{8pt}
    
    \caption{Development set accuracy for the three morphological parsers and a model without morphological parsing. The x-axis tracks the size of the training set. We find that morphological parsing is generally helpful in early rounds, when the training set is very small, but that its importance diminishes as the training set increases. These effects are especially clear for the two MultiNLI-TR dev sets.}
    \label{fig:test_acc}
\end{figure*}

\paragraph{Fine-tuning} We fine-tuned each model on \TurkishNLI\ with the same setting as in \secref{sec:case-study-i}, with the exception that we trained for only 1 epoch. We measured the accuracy on the evaluation sets with an interval of 1,000 training steps to observe the effect of morphological parsing as the dataset grew. \Figref{fig:test_acc} reports the accuracy of all models with respect to fine-tuning steps on \TurkishNLI\ development sets, and \tabref{tab:case_study_ii_results} shows the final accuracies.

\subsubsection{Results} 

\Figref{fig:test_acc} suggests that morphological parsing is beneficial where the training set is small, but its importance largely disappears for large training sets. This is reflected also in the final results in \tabref{tab:case_study_ii_results}. We relate this to the fact that  BERT models create contextual embeddings of both word and subword tokens \cite{kudo-2018-subword, kudo2018sentencepiece, sennrich-etal-2016-neural}. Given a sufficiently large dataset, BERT models can approximate the effects of morphological parsing even for Turkish, a morphologically-rich language.

The trends are not uniform for SNLI-TR and MultiNLI-TR. For SNLI-TR, all three models display a similar learning curve, with a slight edge for Zemberek early on. For MultiNLI-TR, models with morphological parsers are more differentiated. However, all three converge to similar performance at the end of training on both datasets (\Tabref{tab:case_study_ii_results}). 

\begin{table*}[ht]\centering
\setlength{\tabcolsep}{12pt}
\begin{tabular}{l c c c c}
\toprule
\textbf{}  &\multicolumn{2}{c}{\textbf{MultiNLI-TR}} &\multicolumn{2}{c}{\textbf{\FBTR}} \\\cmidrule{2-5}
\textbf{Model Name} & \textbf{XNLI-Dev} & \textbf{XNLI-Test} & \textbf{XNLI-Dev} &\textbf{XNLI-Test} \\\midrule
\textbf{\BertEn} & 66.99\% & 67.74\% & 65.66\% &65.71\% \\
\textbf{\BertMultilingual} & 73.82\% & 72.95\% & 71.61\% & 71.20\% \\
\textbf{BERTurk} &  \textbf{76.75}\% & \textbf{78.72}\% & \textbf{76.43}\% & \textbf{76.43}\% \\
\bottomrule
\end{tabular}
\caption{Accuracy results comparing \TurkishNLI\ with another machine translated dataset. \TurkishNLI\ performed better, but the gap is modest, suggesting that both datasets have value for Turkish NLI. \Figref{fig:xnli_acc} in our supplementary materials provides full learning curves. The results are very similar to those of \tabref{tab:case_study_i_results_table} for MultiNLI, in overall quality and in the ranking of models.}
\label{tab:case_study_iii_results_table}
\end{table*}

In light of these findings, we suggest avoiding the use of morphological parsers for Turkish NLI where the training set is large, since the benefits of such parsers are generally not enough to offset the cost of running them.

\subsection{Case Study III: Evaluating \TurkishNLI\ on Human-Translated Sentences}
\label{sec:case-study-iii}

Thus far, we have used \TurkishNLI\ for both training and assessment. One might worry that machine-translated test sets are not reliable tools for measuring how models will perform on examples written by humans. In this section, we address this concern using the Turkish dev and test portions of XNLI, which were translated entirely by humans. The models we assess on XNLI are those from our first case study as well as models trained on a different machine-translated training dataset, \FBTR. Overall, we find that performance on XNLI is consistently very similar to performance on \TurkishNLI.

\subsubsection{Datasets}

\FBTR\ was created to investigate the performance of cross-lingual sentence embeddings compared to in-language ones (\citealp{conneau-etal-2018-xnli}). It provides machine translations of only the MultiNLI training set, so we report comparisons with just the corresponding section of \TurkishNLI, and we train models only on these two training sets.

\subsubsection{Models}
We used the BERT models from Case Study~I (\secref{sec:case-study-i}) for evaluation. We fine-tuned each model on the training sets of \TurkishMultiNLI\ and \FBTR\ separately, following the same fine-tuning steps as in \secref{sec:case-study-i}, and computed their accuracy on \mbox{XNLI-Dev} and \mbox{XNLI-Test}. 

\subsubsection{Results}

\Tabref{tab:case_study_iii_results_table} provides the results of the experiments. All three models consistently achieve higher accuracy on \mbox{XNLI-Dev} and \mbox{XNLI-Test} when fine-tuned with \TurkishMultiNLI , but the performance difference is modest. \Tabref{tab:case_study_iii_results_table} also illustrates that BERTurk, backed by a Turkish-only training corpus, outperforms the other two models on all eight evaluations. Its performance is followed by \BertMultilingual, which is trained on a corpus with texts in multiple languages, including Turkish. The same result was also shown in Case Study~I using the evaluation splits of \TurkishNLI. Therefore, machine-translated \TurkishMultiNLI\ and human-translated XNLI display similar characteristics across evaluations, which lends further credence to our claim that MT can help provide a viable path to robust Turkish NLI. 

To better understand how in-language pretraining (BERTurk) helps, we investigated the 57 hypotheses from XNLI-Dev and XNLI-test where  BERTurk was successful and \BertMultilingual\ was not. For these sentences, we observed that the \BertMultilingual\ tokenizer was often unable to segment the words into meaningful Turkish subword units, most likely due to its training on a multilingual corpus. For instance, \BertMultilingual\ often could not segment the suffix \mbox{``-me/ma''}, which negates a verb in Turkish, and thus bears crucial semantics for many contradiction examples \citep{gururangan-etal-2018-annotation}. This shows that in-language training is essential not only for good vector representations but also for effective tokenization.

We also hypothesize that subtle lexical distinctions are another factor in the performance difference between BERTurk and \BertMultilingual. For example, though \BertMultilingual\ successfully identified the semantic relations created by frequent pairs such as ``hi\c{c}'' (`any') and ``hepsi'' (`all'), it missed many other distinctions like these. We propose that this is due to the more limited vocabulary of \BertMultilingual\ for Turkish and the more robust word representations in BERTurk.

In addition to manual inspection, we computationally analyzed the pairs where \BertMultilingual\ was unsuccessful and BERTurk was successful. We computed the frequency of each semantic class in the \BertMultilingual\  predictions for these sentences and observed that the neutral class is the most common. This perhaps reflects the fact that neutral is the default choice where the model cannot robustly identify a semantic relation.

\section{Conclusion} 

We created and released the first large Turkish NLI dataset, \TurkishNLI, by machine translating SNLI and MultiNLI. Though English and Turkish have very different grammars and thus stress-test automatic approaches, our team of experts judged the translations to be of very high quality and to preserve the original NLI labels consistently. These results suggest that MT can help address the paucity of datasets for Turkish NLI. We release code, models, and data publicly for further research.

We also used \TurkishNLI\ to investigate central issues in Turkish NLI. First, we used \TurkishNLI\ to analyze the effects of in-language pretraining. Second, we compared three morphological parsers for Turkish with simpler tokenization schemes. We found that a Turkish-only pretraining regime can enhance Turkish models significantly, and that morphological parsing is arguably worth its costs only when the training dataset is small. In our final case study, we returned to the general issue of translation quality, but now from the perspective of developing NLI systems. We showed that models trained on \TurkishMultiNLI\ perform well on the expert-translated test set from XNLI.

On the basis of these findings, we argue that MT can be more widely adopted for advancing NLP studies on resource-constrained languages. Though language-dependent tasks like dependency parsing are challenging to translate, MT can efficiently transfer large and expensive-to-create labeled datasets from English to other languages in many NLP tasks, including text classification, question answering, and text summarization. In addition, MT will presumably get cheaper, faster, and better over time, thereby further strengthening our core claims.

\section*{Acknowledgments}
This research was supported by the AWS Cloud Credits for Research Program (formerly AWS Research Grants). E.Budur is thankful for the support provided by The Scientific and Technological Research Council of Turkey (T\"{U}B\.{I}TAK) and Council of Higher Education (Y\"{O}K) under B\.{I}DEB 2214/A and 100/2000 graduate research scholarship programs, respectively.  R.\"{O}z\c{c}elik  gratefully acknowledges the graduate research scholarship by T\"{U}B\.{I}TAK under B\.{I}DEB 2211/A program. 

The authors gratefully acknowledge that the computational parts of this study has been mostly performed at Bo\u{g}azi\c{c}i \mbox{TETAM DGX-1} GPU Cluster and partially carried out at T\"{U}B\.{I}TAK ULAKB\.{I}M High Performance and Grid Computing Center (TRUBA resources) and Stanford Research Computing Center (FarmShare).

We thank Alara Dirik, Almira Ba\u{g}lar, Berfu B\"{u}y\"{u}k\"{o}z, Berna Erden, Fatih Mehmet G\"{u}ler, G\"{o}k\c{c}e Uludo\u{g}an,  G\"{o}zde Aslanta\c{s}, Havva Y\"{u}ksel, Melih Barsbey, Melike Esma \.{I}lter, Murat Karademir, Ramazan Pala, Selen Parlar, Tu\u{g}\c{c}e Ulutu\u{g}, Utku Yavuz for their annotation support and vital contributions. We are grateful also to Stefan Schweter and Kemal Oflazer for sharing the dataset that BERTurk was trained on and Omar Khattab, Dallas Card, Yiwei Luo, and many more researchers including the anonymous reviewers for their valuable advice, discussion and insightful comments.

\bibliography{bibliographies/anthology, bibliographies/emnlp2020, bibliographies/manual_entries}
\bibliographystyle{acl_natbib}

\clearpage
\appendix
\label{sec:appendix}
\begin{table*}[b]
\centering
\begin{tabular}{llcc}
\large{\textbf{Appendices}}&&& \\
& & \textbf{English} & \textbf{Turkish} \\ \hline

\multicolumn{1}{|l|}{}  & \multicolumn{1}{p{2.5cm}|}{\textbf{Premise}} & \multicolumn{1}{p{5cm}|}{Several people are on stage preparing for a show.} & \multicolumn{1}{p{5cm}|}{Birka\c{c} ki\c{s}i sahnede g\"{o}steri i\c{c}in haz{\i}rlan{\i}yor.} \\ \cline{2-4}

\multicolumn{1}{|l|}{} & \multicolumn{1}{p{2.5cm}|}{ \textbf{Entailment}} & \multicolumn{1}{p{5cm}|}{People are setting up for a show.} & \multicolumn{1}{p{5cm}|}{İnsanlar bir g\"{o}steri i\c{c}in haz{\i}rlan{\i}yor.} \\ \cline{2-4}

\multicolumn{1}{|l|}{} & \multicolumn{1}{p{2.5cm}|}{ \textbf{Contradiction}} & \multicolumn{1}{p{5cm}|}{A house is being demolished.} & \multicolumn{1}{p{5cm}|}{Bir ev y{\i}k{\i}l{\i}yor.}                                                                              \\ \cline{2-4} 
\multicolumn{1}{|l|}{\multirow{-4}{*}{\textbf{SNLI}}}  & \multicolumn{1}{p{2.5cm}|}{ \textbf{Neutral}} & \multicolumn{1}{p{5cm}|}{A crew is getting ready for a rock concert.} & \multicolumn{1}{p{5cm}|}{Bir ekip rock konseri i\c{c}in haz{\i}rlan{\i}yor.} \\ \hline

\multicolumn{1}{|l|}{} & \multicolumn{1}{{p{2.5cm}|}}{\textbf{Premise}} & \multicolumn{1}{p{5cm}|}{ All rooms have color TV, alarm clock/radio, en-suite bathrooms, real hangers, and shower massage.} & \multicolumn{1}{p{5cm}|}{T\"{u}m odalarda renkli TV, \c{c}alar saat/radyo, en-suite banyo, ger\c{c}ek ask{\i}lar ve du\c{s} masaj{\i} vard{\i}r.} \\ \cline{2-4} 

\multicolumn{1}{|l|}{} & \multicolumn{1}{p{2.5cm}|}{\textbf{Entailment}} & \multicolumn{1}{p{5cm}|}{ All rooms also contain a ceiling fan and outlets for electronics.} & \multicolumn{1}{p{5cm}|}{ T\"{u}m odalarda ayr{\i}ca tavan vantilat\"{o}r\"{u} ve elektronik prizler bulunmaktad{\i}r.} \\ \cline{2-4} 

\multicolumn{1}{|l|}{} & \multicolumn{1}{p{2.5cm}|}{ \textbf{Contradiction}} & \multicolumn{1}{p{5cm}|}{ You will not find a TV or alarm clock in any of the rooms.} & \multicolumn{1}{p{5cm}|}{ Odalar{\i}n hi\c{c}birinde TV veya \c{c}alar saat bulunmamaktad{\i}r.} \\ \cline{2-4} 

\multicolumn{1}{|l|}{\multirow{-8}{*}{\textbf{MultiNLI}}} & \multicolumn{1}{p{2.5cm}|}{ \textbf{Neutral}}       & \multicolumn{1}{p{5cm}|}{Color TVs, alarms, and hangers can be found in all rooms.} & \multicolumn{1}{p{5cm}|}{ T\"{u}m odalarda renkli TV'ler, alarmlar ve ask{\i}lar bulunur.} \\ \hline
\end{tabular}

\caption{Sample translations from SNLI and MultiNLI into \TurkishNLI. Each premise is associated with a hypothesis from each of the three NLI categories.}
\label{tab:sample_translations_table_all}
\end{table*}
\begin{table*}[!ht]\centering
\setlength{\tabcolsep}{12pt}
\begin{tabular}{l c c c c}
\toprule
\textbf{}  &\textbf{SNLI} &\multicolumn{2}{c}{\textbf{MultiNLI}} \\\cmidrule{2-4}
\textbf{Model Name} & \textbf{Test} & \textbf{Matched Dev} &\textbf{Mismatched Dev} \\\midrule
\textbf{\BertEn\ } & \textbf{90.13\%} & \textbf{83.16\%} &  \textbf{83.95\%} \\

\textbf{\BertMultilingual\ } &89.02\% & 81.74\% & 82.13\% \\
\textbf{BERTurk } &85.84\% & 75.16\% & 75.60\% \\
\bottomrule
\end{tabular}
\caption{Accuracy of the cased models in \Tabref{tab:case_study_i_results_table} trained on SNLI and MultiNLI. We used the same fine-tuning and evaluation procedures. \BertEn\ ranked the first and \BertMultilingual\ ranked the second, emphasizing the importance of in-language training one-more time as in \Secref{sec:case-study-i}.}.
\label{tab:case_study_i_app_table}
\end{table*}

\begin{figure*}[tp]
    \centering
    \includegraphics[width=\textwidth]{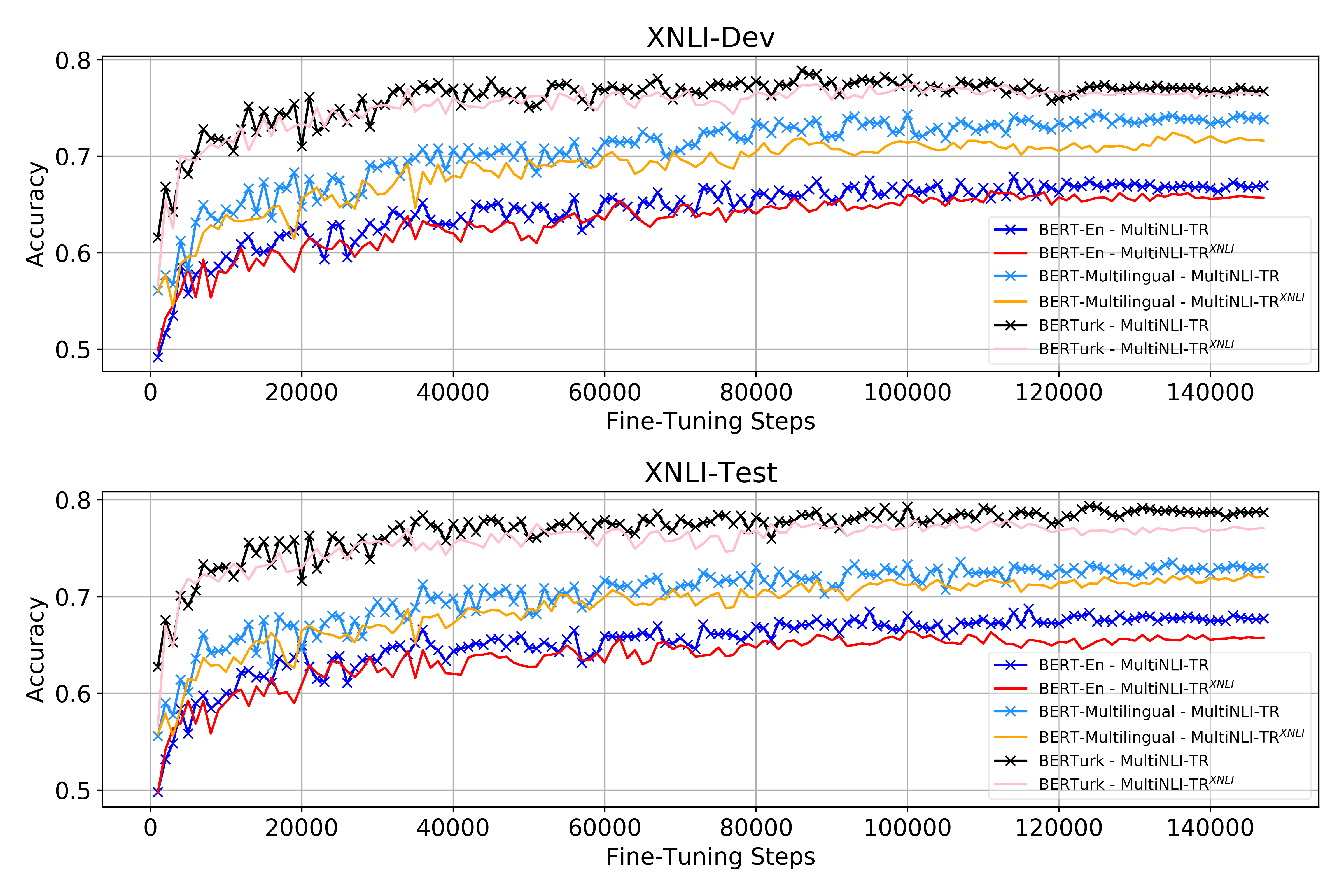}
    
    \vspace{8pt}
    \caption{XNLI-Dev and XNLI-Test accuracy of three transformer models trained on \TurkishMultiNLI\ and \FBTR. The x-axis tracks the training set size. We find that models trained on \TurkishMultiNLI\ are superior to their \FBTR\ counterparts from the start of the training until the end.}    \label{fig:xnli_acc}
\end{figure*}

\begin{table*}[ht]\centering
\setlength{\tabcolsep}{12pt}
\begin{tabular}{l c c c c}
\toprule
\textbf{}  &\multicolumn{2}{c}{\textbf{MultiNLI-TR}} &\multicolumn{2}{c}{\textbf{\FBTR}} \\\cmidrule{2-5}
\textbf{Model Name} & \textbf{XNLI-Dev-TR} & \textbf{XNLI-Test-TR} & \textbf{XNLI-Dev-TR} &\textbf{XNLI-Test-TR} \\\midrule
\textbf{\BertEn} & 70.11\% & 70.11\% & 68.42\% & 67.70\% \\

\textbf{\BertMultilingual} & 75.85\% & 74.79\% & 74.69\% & 73.77\% \\

\textbf{BERTurk} &  \textbf{80.11}\% & \textbf{79.52}\% & \textbf{79.95}\% & \textbf{78.40}\% \\
\bottomrule
\end{tabular}
\caption{Accuracy results of the models in \Tabref{tab:case_study_iii_results_table} for machine translated XNLI. The outcomes agree with the ones in \Secref{sec:case-study-iii}, suggesting that machine translated sentences can be used to evaluate Turkish NLI models. Here we note that, \mbox{XNLI-Dev-TR}, \mbox{XNLI-Test-TR} and \TurkishMultiNLI\ are translated with the same MT service, whereas \FBTR\ used a different one. Though this might result in a positive bias for \TurkishMultiNLI\ models, we report the accuracy of \FBTR\ models as well for the sake of completeness.}
\label{tab:case_study_iii_app_table}
\end{table*}

\begin{figure*}[tp]
    \centering
    \includegraphics[width=0.65\textwidth]{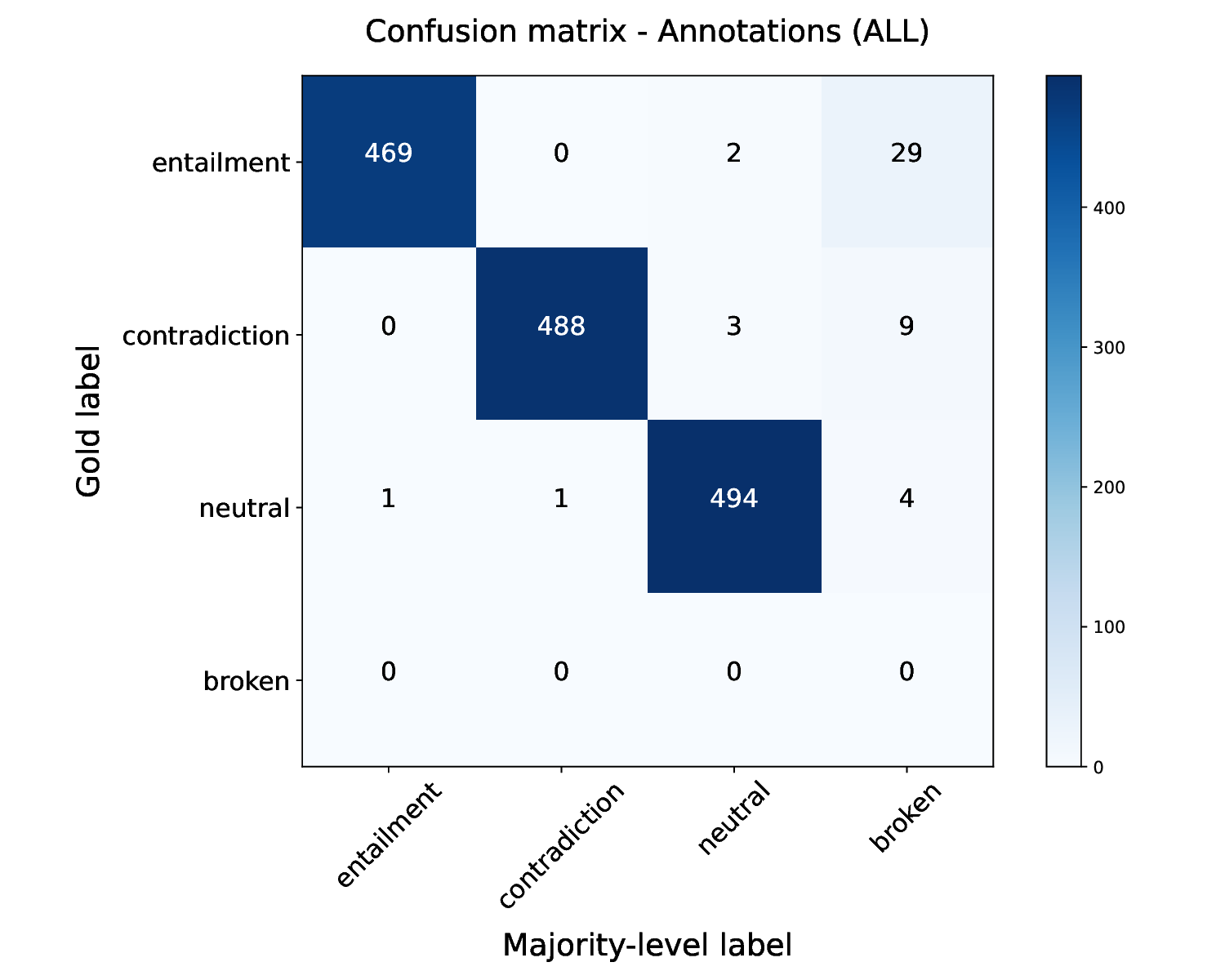}
    
    \vspace{8pt}
    \caption{Confusion matrix for the majority-level label consistency results of all annotations in Table \ref{tab:dataset_quality_stats_v2}.  The label ``broken'' corresponds to the pairs which have either major translation error or no  majority-level label.}    \label{fig:confusion_matrix_annotation}
\end{figure*}

\end{document}